# QuickNet: Maximizing Efficiency and Efficacy in Deep Architectures

Author: Tapabrata Ghosh[1] (admin@Ingemini.com)
Ingemini LLC[1]

Abstract:

    We present QuickNet, a fast and accurate network architecture that is both faster and significantly more accurate than other "fast" deep architectures like SqueezeNet. Furthermore, it uses less parameters than previous networks, making it more memory efficient. We do this by making two major modifications to the reference "Darknet" model (Redmon et al, 2015): 1) The use of depthwise separable convolutions and 2) The use of parametric rectified linear units. We make the observation that parametric rectified linear units are computationally equivalent to leaky rectified linear units at test time and the observation that separable convolutions can be interpreted as a compressed Inception network (Chollet, 2016). Using these observations, we derive a network architecture, which we call QuickNet, that is both faster and more accurate than previous models. Our architecture provides at least four major advantages: (1) A smaller model size, which is more tenable on memory constrained systems; (2) A significantly faster network which is more tenable on computationally constrained systems; (3) A high accuracy of 95.7% on the CIFAR-10 Dataset which outperforms all but one result published so far, although we note that our works are orthogonal approaches and can be combined (4) Orthogonality to previous model compression approaches allowing for further speed gains to be realized.

# 1 Introduction and Motivation

### 1.1 Introduction

    Convolutional neural networks have quickly become the state of the art in image classification, object recognition and other previously difficult computer vision tasks. In fact, it can be argued that convolutional neural networks alone are responsible for the newfound ability of perception for computers. In the journey towards artificial intelligence, they will have a key role.

    Convolutional neural networks were first introduced several years ago with LeNet, however, in recent years, architectural improvements, new activation functions allowing gradient flow, increased computational ability and massive datasets have allowed them to progress into the mainstream. Furthermore, the revolution of depth has been a vital component, and in fact, for a time, the rule seemed to be that the deeper the network is, the

better its accuracy will be. The initial LeNet contained only five layers, then came AlexNet with 12 layers, then VGG with 19 layers, then ResNet with all the way up to 1001 layers. These were all trained with the use of massively parallel GPUs which enabled training and inference of extremely deep networks, delivering phenomenal accuracy compared to previous computer vision models that relied upon hand engineering features and support vector machines.

## 1.2 Motivation

However, the aforementioned increased computational ability is not available in all applications and on all platforms. In particular, the smartphone has become a global phenomenon with far reaching social impact, yet the tremendous required computation for modern deep networks is impossible to achieve on a smartphone. In addition, other applications that would greatly benefit from deep learning, such as autonomous vehicles are unable to do so due to the computational constraints. However, making small architectures that are unable to match the accuracy levels of their full-fledged counterparts often negates the advantage of leveraging deep learning in the first place. Therefore, we assert that in order to provide an optimal benefit, a balance between computational tractability and accuracy preservation must be made. To that regard, we focus on shrinking the computational complexity of the network while still preserving the accuracy of the network. To tackle the problem, we deliver QuickNet, a faster and more accurate network architecture.

# 2 Related Work

## 2.1 Darknet, SqueezeNet and other "Fast" Architectures

Our work builds heavily upon previous works and combines several previous ideas in a novel way. Redmon et al (2015) provide a baseline network termed "Darknet" which is used in the YOLO object detection framework for real-time performance. However, this was intended for real-time performance on an Nvidia Titan X GPU which has ~10 TFLOPs of compute power. Nonetheless, this network provides a good baseline which we heavily modify in order to improve accuracy and runtime.

Iandola et al (2016) proposed SqueezeNet, a fast and memory lightweight architecture that matched AlexNet accuracy. However, in modern times, AlexNet, and subsequently, SqueezeNet has a relatively massive error rate of 18.04%, which is not nearly as useful as it seems at first. Therefore, we propose an architecture that is both faster *and* more accurate than existing state of the art models.

## 2.2 Model Compression

Much work has been done on model compression, especially pruning (Han et al, 2015) and quantization (Vanhoucke et al). The combination of the two resulted in a generic pipeline to compress a neural network and its weights termed "Deep Compression" (Han et al, 2015)

## 2.3 Depthwise Separable Convolutions and Parametric Rectified Linear Units

This work heavily relies upon depthwise separable convolutions and parametric rectified linear units (PReLUs) in order to provide the performance and accuracy. Depthwise separable convolutions were introduced and utilized by Chollet, 2016 in the XCeption architecture although they were used previously as well. Depthwise separable convolutions more effectively capture channel-wise features by first performing a depthwise convolution and then performing a pointwise (or 1x1) convolution. Parametric rectified linear units (PReLUs) were introduced by He et al in 2015 in order to improve performance on the Imagenet (Russakovsky et al, 2014) dataset. PReLUs introduce another parameter to take the place of the slope of the negative part of the leaky ReLU.

# 3 The QuickNet Architecture

## 3.1 Memory and Energy Conscious Design

It has been shown that memory access, not compute is the primary source of power consumption in deep neural networks. Han et al (2016) showed that for a commercial 40nm process, DRAM access is more than 200 times as energy intensive than a 32 bit multiply. Therefore, it appears that minimizing parameters is the clear path towards an energy efficient and lightweight neural architecture. However, in addition to parameter count, activation maps in intermediate layers also require large amounts of memory store, even at inference time. Nevertheless, the QuickNet architecture also minimizes the parameter count, with less than 14.24 megabytes at 32 bits (3.56 million parameters), and down to less than a megabyte with the use of Deep Compression, (Han et al, 2015). We observe that a compression rate of 15x from Deep Compression is more reasonable than the reported 50x.

The issue with focusing upon minimizing parameter count is that the Deep Compression pipeline (Han et al, 2015) already provides a simple and effective pipeline to reduce model size by 50x through a combination of pruning, quantization and Huffman coding with no accuracy loss. This in and of itself is sufficient to reduce the model size to a tractable amount. By contrast, work on taming the computational complexity of deep neural networks while still maintaining good accuracy has been lagging behind. Consequently, we focus upon reducing the computational complexity of our network and maintaining accuracy and make the observation that compression already is well researched and stable.

## 3.2 Computationally Conscious Design

Due to the aforementioned reasons, we choose to focus on creating a computationally efficient architecture while still maintaining high accuracy, To this extent, we introduce the QuickNet Architecture.

## 3.3 Depthwise Separable Convolutions

Chollet et al utilized depthwise separable convolutions in 2016 which are faster and use less parameters due to the preservation of cross-channel features. We also utilize depthwise separable convolutions in our architecture. As a result, we have far less computation (separable convolutions are faster) and use less parameters (since separable convolutions allow greater expressivity of channelwise features). A key difference between the XCeption architecture and QuickNet is the removal of residual connections in the interest of preserving computational tractability since residual connections have a massive memory impact in terms of storing activation maps. Chollet et al also observed that separable convolutions can be interpreted as an "extreme" Inception (Szegedy et al) block and indeed, this interpretation is supported by our findings. In this light, QuickNet can be regarded as a compressed yet still improved Inception network. We use 3x3 separable convolutions in order to maximize accuracy, reduce computation and reduce memory size.

## 3.4 Parametric Rectified Linear Units

He et al introduced parametric rectified linear units in order to improve performance on the imagenet dataset in 2015. We make the observation that at inference time, the PReLU is computationally equivalent to the Leaky ReLU with a minimal parameter increase, yet provides a noticeable performance increase. To the best of our knowledge, this is the first time this observation has been made and leveraged in a speed-oriented deep learning architecture.

## 3.5 Global Average Pooling Instead of Fully-Connected Layers

We utilize a simple and effective trick to massively reduce the parameter count and computational cost by replacing the fully connected layers at the end of the network (before feeding into the softmax regression) with global average pooling layers. We note that spatial pyramidal pooling could be used instead as proposed by He et al (2016) and would provide greater accuracy, but we choose to use global average pooling instead for greater inference speed.

## 3.6 An Efficient and Effective Entry Stem

In order to reduce the activation map sizes and computation costs, we replace the traditional entry flow stem of several layers with a simple 5x5 convolution layer with no padding for a simple and efficient entry flow that actually *increases* accuracy at the start of the network, before feeding into the repeating blocks.

## 3.7 Putting it All Together: QuickNet

We combine these components to build the QuickNet macroarchitecture, which follows a quasi-repetitive and "Inception-esque" construction, consisting of an entry flow, quasi-repetitive blocks of 3x3 separable convolutions (quasi-repetitive since the number of filters at each layer doubles every N layers) and an ending global average pooling. A full visualization is provided in the appendix and a higher-resolution visualization can be found here: https://drive.google.com/open?id=0BzEsy1iWjGixOGxnbUtYMElaZGs

# 4 Experimental Results

The QuickNet network was trained on CIFAR-10 with data augmentation in the Keras framework (Chollet, 2015). We used dropout (Srivastava et al, 2014) with a value of 0.5 and batch normalization. We used the standard cross-entropy loss as a loss function. We used a validation/test set that was 6000 images (10%) randomly chosen which the network never looked at. This validation/test set was used to measure the accuracy.

| Network Architecture | CIFAR-10 Test Error (in %) |
|---|---|
| AlexNet | 18.04 |
| Network in Network | 8.81 |
| Maxout | 9.38 |
| VGG-16 | 7.55 |
| All-Convolutional | 7.25 |
| Fractional Max Pooling | **3.47** |
| QuickNet (this paper) | 4.3 |

With early stopping at epoch 190, our network achieves an accuracy of 95.7% or an error rate of 4.3%. To the best of our knowledge, our results on the CIFAR-10 Dataset are surpassed only by Fractional Max Pooling (Graham, 2014). Furthermore, our approach is orthogonal to Fractional Max Pooling and could be combined, but we decided to forgo it in order to (1) Preserve computational tractability since all available implementations of Fractional Max Pooling take significantly longer (up to 15x in Lasagne) than traditional pooling methods with no clear path towards optimization and (2) Preserve memory tractability since memory access is the most energy intensive operation and would massively increase with the use of Fractional Max Pooling. We note in our future work that this may be a possible avenue to improve accuracy and are working with reconciling it with the computational limits of the target platform.

An additional observed characteristic of QuickNet that is of note is the relatively fast convergence of the architecture (80% accuracy within 70 epochs), which may one day open the door to local training and updating of networks.

Note that the reported results are the best of two runs (both within ~0.5% of each other), a more exhaustive hyperparameter search and experimentation with different adaptive optimizers such as Adam or RMSProp may yield even better results.

# 5 Computational Performance

Our model achieves 15 frames per second on a low-power CPU which is already far more practically applicable than other state of the art models. However, we posit that massive gains be be obviously realized with the 8-bit quantization proposed by Vanhoucke et al, this could potentially run at 60fps on low power CPUs, which is considered the threshold for real-time capabilities. This opens the door to state of the art performance deep architectures to be deployed on smartphones.

Furthermore, as mentioned previously, our model is compatible with the Deep Compression pipeline by Han et al, allowing us to easily reduce the model size by more than an order of magnitude, making QuickNet tractable in memory constrained platforms.

# 6 Conclusion

In this paper, we called for the need for a computationally efficient and accurate neural network architecture by observing and noting that a generic and stable memory compression pipeline already exists, obviating the need to focus on reducing parameter count when designing deep architectures. We also noted that to date, computationally efficient deep architectures have lagged behind, with most works in literature incorrectly assuming that a lower parameter count automatically translates to a computationally efficient architecture.

To that end, we have presented QuickNet, an architecture that achieves phenomenal performance and still manages to preserve computational tractability. In addition, we note that QuickNet is an orthogonal approach that is still tenable to previous model compression pipelines, most notably, Deep Compression. Finally, we note some potential improvements and provide early steps and early experimentations towards improving upon the QuickNet baseline further. QuickNet achieves a 95.7% accuracy rate on CIFAR-10, the second best to date and runs at 15 frames per second even on low-power CPUs. We also show a clear pathway to achieve 60 frames per second on low-power CPUs, enabling smartphone deployment.

Although in this architecture we have focused upon image classification, we note that architectures for image classification have had no trouble with domain adaptation to other applications, such as object detection, semantic regression and others. We hope that QuickNet can become a baseline model used for applications where both computational tractability and accuracy are desired and required.

# 7 Future Work and Architecture Manifold Exploration

Our findings and observations indicate that there is a *manifold* of "accurate architectures" as opposed to a *set* of "accurate architectures". This assertion is supported by literature observations as well as our own experiments showing how permutations of the QuickNet macro-architecture can achieve similar accuracy levels. This ties into the macroarchitecture and microarchitecture taxonomy of deep learning architectures.

In future work, we hope to explore the impact of residual connections and dense connections upon accuracy. Although the latter will (significantly) negatively impact computational and memory tractability, from a pure accuracy standpoint, the combination is of utmost interest. The former is far more computationally tractable and experiments exploring the impact of residual connections were promising but were aborted due to computational and temporal constraints.

In order to push even higher accuracy in future works, we would utilize mean-only Batch Normalization and Weight Normalization which should actually improve inference runtimes. Preliminary experiments along this path were promising but were aborted due to computational and temporal constraints.

Finally, in the interest of maximizing accuracy with no regard to computational cost, we would like to explore a QuickNet architecture that utilizes mean-only Batch Normalization and Weight Normalization, Spatial Pyramidal Pooling, densely connected blocks, Spatial Transformer Layers and Group-Convolutions. No experiments were attempted by us along this axis in this work, although we intend to investigate this path in future works.

# References/Bibliography:

Graham, Benjamin. Fractional max-pooling. CoRR, abs/1412.6071, 2014.
    URL http://arxiv.org/ abs/1412.6071.

Zagoruyko, S. (n.d.). 92.45% on CIFAR-10 in Torch.
    Retrieved January 08, 2017, from http://torch.ch/blog/2015/07/30/cifar.html

P. (2017, January 04). Pjreddie/darknet. Retrieved January 08, 2017,
    from https://github.com/pjreddie/darknet

Joseph Redmon and (2015). You Only Look Once: Unified, Real-Time Object Detection.
    *CoRR, abs/1506.02640*, .

Francois Chollet (2016). Xception: Deep Learning with Depthwise Separable
    Convolutions. *CoRR, abs/1610.02357*, .

Vanhoucke, V., Senior, A., and Mao, M. Z. (2011). Improving the speed of neural networks
    on cpus. In Proc. Deep Learning and Unsupervised Feature Learning
    NIPS Workshop.

Forrest N. Iandola and (2016). SqueezeNet: AlexNet-level accuracy with 50x
 fewer parameters and <0.5MB Model Size. *CoRR, abs/1602.07360*, .

C. Szegedy, W. Liu, Y. Jia, P. Sermanet, S. Reed, D. Anguelov, D. Erhan, V. Vanhoucke, and A. Rabinovich. Going deeper with convolutions. In Proceedings of the IEEE Conference on Computer Vision and Pattern Recognition, pages 1–9, 2015.

C. Szegedy, V. Vanhoucke, S. Ioffe, J. Shlens, and Z. Wojna. Rethinking the inception architecture for computer vision. arXiv preprint arXiv:1512.00567, 2015.

M. Wang, B. Liu, and H. Foroosh. Factorized convolutional neural networks. arXiv preprint arXiv:1608.04337, 2016.

O. Russakovsky, J. Deng, H. Su, J. Krause, S. Satheesh, S. Ma, Z. Huang, A. Karpathy, A. Khosla, M. Bernstein, et al. Imagenet large scale visual recognition challenge. 2014

F. Chollet. Keras. https://github.com/fchollet/keras, 2015

A. Howard. Mobilenets: Efficient convolutional neural networks for mobile vision applications. Forthcoming.

S. Han, H. Mao, and W. Dally. Deep compression: Compressing DNNs with pruning, trained quantization and huffman coding. arxiv:1510.00149v3, 2015a.

Song Han, Xingyu Liu, Huizi Mao, Jing Pu, Ardavan Pedram, Mark A Horowitz, and William J Dally. Eie: Efficient inference engine on compressed deep neural network. International Symposium on Computer Architecture (ISCA), 2016a.

K. He, X. Zhang, S. Ren, and J. Sun. Delving deep into rectifiers: Surpassing human-level performance on imagenet classification. In ICCV, 2015a.

Y. LeCun, B.Boser, J.S. Denker, D. Henderson, R.E. Howard, W. Hubbard, and L.D. Jackel. Backpropagation applied to handwritten zip code recognition. Neural Computation, 1989.

Min Lin, Qiang Chen, and Shuicheng Yan. Network in network. arXiv:1312.4400, 2013.

Vinod Nair and Geoffrey E. Hinton. Rectified linear units improve restricted boltzmann machines. In ICML, 2010.

Nitish Srivastava, Geoffrey Hinton, Alex Krizhevsky, Ilya Sutskever, and Ruslan Salakhutdinov. Dropout: a simple way to prevent neural networks from overfitting. JMLR, 2014.

Ian J. Goodfellow, David Warde-Farley, Mehdi Mirza, Aaron Courville, Yoshua Bengio
Maxout Networks, arXiv:1302.4389 2013